\documentclass[10pt,twocolumn,letterpaper]{article}

\usepackage[pagenumbers]{cvpr} 

\usepackage{graphicx}
\usepackage{amsmath}
\usepackage{amssymb}
\usepackage{booktabs}
\usepackage{tabularray}
\newcommand{\proportion}{1}

\newcommand{\generictext}{\textit{In this example, we compare various raw-to-raw transformation methods across different smartphones and methods. Due to variations in Image Signal Processing (ISP) pipelines, including the use of distinct Chromatic Adaptation Transforms (CATs) and 3D Look-Up Tables (3D-LUTs), the final processed images differ significantly between devices. However, our analysis focuses solely on evaluating the consistency of raw-to-raw transformation methods, independent of the specific ISP outputs.}\\}

\definecolor{cvprblue}{rgb}{0.21,0.49,0.74}
\usepackage[symbol]{footmisc}
\usepackage[pagebackref,breaklinks,colorlinks,allcolors=cvprblue]{hyperref}

\usepackage[capitalize]{cleveref}
\crefname{section}{Sec.}{Secs.}
\Crefname{section}{Section}{Sections}
\Crefname{table}{Table}{Tables}
\crefname{table}{Tab.}{Tabs.}

\usepackage{tabularx} 
\newcolumntype{L}{>{\raggedright\arraybackslash}X}
\newcolumntype{R}{>{\raggedleft\arraybackslash}X}
\newcolumntype{C}{>{\centering\arraybackslash}X} 

\usepackage{color, colortbl} 
\definecolor{LightGray}{rgb}{0.9,0.9,0.9}
\definecolor{cvprblue}{rgb}{0.21,0.49,0.74}

\usepackage{multirow}

\usepackage{graphicx} 
\usepackage{animate}

\usepackage{pifont} 
\newcommand*\colourcheck[1]{%
  \expandafter\newcommand\csname #1check\endcsname{\textcolor{#1}{\ding{52}}}%
}
\newcommand*\colourcross[1]{%
  \expandafter\newcommand\csname #1cross\endcsname{\textcolor{#1}{\ding{56}}}%
}

\colourcheck{green}
\colourcheck{yellow}
\colourcross{red}

\usepackage{graphbox}

\usepackage{enumitem}
\setitemize{noitemsep,left=0pt, topsep=0pt,parsep=0pt,partopsep=0pt}

\makeatletter
\let\@oldmaketitle\@maketitle
\renewcommand{\@maketitle}{\@oldmaketitle
\centering
\vspace{-0.3cm}
\includegraphics[width=\textwidth]{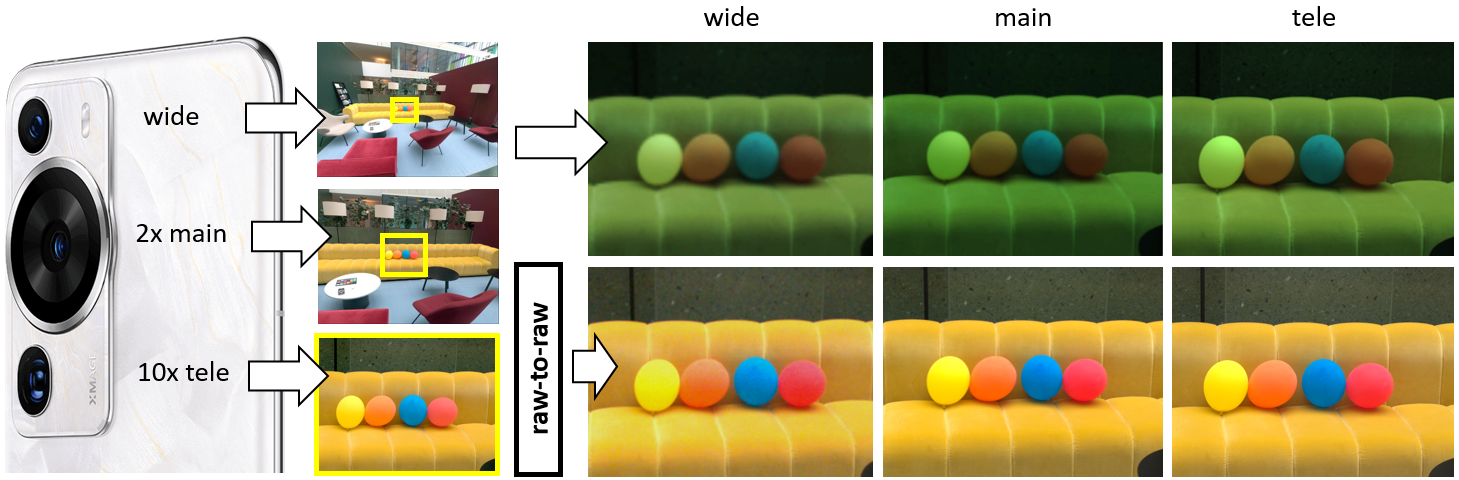}

\captionof{figure}{
\textbf{Hybrid zoom with our physically informed raw-to-raw method}.
The top row shows the raw RGB-space images by different camera sensors. The bottom row displays images transformed using our proposed physically informed raw-to-raw method to match the main camera's color space, followed by a simplified color post-processing pipeline. Our method effectively reduces color discrepancies across images captured by the wide, main, and telephoto cameras, enabling a seamless hybrid zoom experience. Notably, this approach works well despite the fact that the cameras are not active simultaneously and have different fields of view. \vspace{0.3cm}
}\label{fig:eyecatcher}
}
\makeatother

\def\paperID{12976} 
\def\confName{CVPR}
\def\confYear{2025}

\newcommand{\methodname}{NPM} 

\title{Beyond Calibration: Physically Informed Learning for Raw-to-Raw Mapping}

\author{Peter Grönquist\footnote[1]{These authors contributed equally to this work.}\\
Huawei Research Zürich\\
Zürich, Switzerland\\
{\tt\small peter.groenquist@huawei.com}
\and
Stepan Tulyakov\footnote[1]{These authors contributed equally to this work.}\\
Huawei Research Zürich\\
Zürich, Switzerland\\
{\tt\small stepan.tulyakov@huawei.com}
\and
Dengxin Dai\\
Huawei Research Zürich\\
Zürich, Switzerland\\
{\tt\small dengxin.dai@huawei.com}
}
\begin{document}
\maketitle

\begin{abstract}

Achieving consistent color reproduction across multiple cameras is essential for seamless image fusion and Image Processing Pipeline (ISP) compatibility in modern devices, but it is a challenging task due to variations in sensors and optics. Existing raw-to-raw conversion methods face limitations such as poor adaptability to changing illumination, high computational costs, or impractical requirements such as simultaneous camera operation and overlapping fields-of-view. We introduce the Neural Physical Model (NPM), a lightweight, physically-informed approach that simulates raw images under specified illumination to estimate transformations between devices. The NPM effectively adapts to varying illumination conditions, can be initialized with physical measurements, and supports training with or without paired data. Experiments on public datasets like NUS and BeyondRGB demonstrate that NPM outperforms recent state-of-the-art methods, providing robust chromatic consistency across different sensors and optical systems.
\end{abstract}

\section{Introduction}
\label{sec:intro}

Modern devices like smartphones, drones, virtual reality headsets, and security systems increasingly rely on multiple cameras working together to deliver a unified visual experience. For instance, smartphones combine images from different lenses to enable a smooth zoom functionality~\cite{wu2024}. However, achieving seamless image fusion requires consistent chromatic properties across all cameras, which is challenging due to variations in optics, coatings, and sensor technologies.

\footnotetext[1]{These authors contributed equally to this work.}

Additionally, image signal processing (ISP) pipelines and algorithms are typically fine-tuned for specific camera characteristics. However, during the manufacturing process, individual cameras and their parts can be provided by different suppliers, or have different adjustments, to optimize for cost and quality. This often results in inconsistent color reproduction and sub-optimal algorithm performance, highlighting the need for robust raw-to-raw conversion methods to standardize raw images across different sensors and optical systems. These methods can be loosely classified as belonging to one of three categories: calibration-based, translation-based, and photometric alignment. 

\textit{Calibration-based methods}~\cite{ilie2005, joshi2005, sabater2017, nguyen2014, wilburn2005} aim to linearize and match camera responses to a set of reference responses on selected colors, such as the Macbeth ColorChecker. These methods are simple and fast but perform poorly under varying illumination as noted in ~\cite{nguyen2014}.

\textit{Translation-based methods}~\cite{afifi2021, perevozchikov2024}, akin to image-to-image translation techniques~\cite{zhao2020, zhu2017, park2020}, use convolutional neural networks to transform raw images from one camera to match the characteristics of another. This approach performs an implicit estimation of the spectral power distribution of the illumination from RGB data, which is an inherently ill-posed problem. Although these methods typically outperform calibration-based techniques, they rely on computationally intensive and parameter-heavy auto-encoder networks, requiring a lot of training data and treating the transformation as a black-box process without any physical basis.    

\textit{Photometric alignment} methods~\cite{song2022}, similar to calibration-based methods, adjust the colors in an image to match, but they do this by using the colors of shared features in a reference image as targets. These methods can potentially offer the best chromatic consistency; however, they require a robust feature matching process and the operation of two cameras simultaneously, which leads to additional energy consumption and computational demand. Moreover, these methods are not applicable when cameras have a non-overlapping field of view, such as the front and back cameras of a smartphone, or when cameras are not present on the device simultaneously, as is the case when integrating a new sensor into a device with a fixed image signal processing pipeline.\\

To address the limitations of previous methods, we propose the \textbf{Neural Physical Model~(\methodname{})} a new raw-to-raw conversion approach with several key advantages. Unlike the calibration-based methods, the proposed approach effectively adapts to varying illumination conditions. In contrast to the photometric alignment methods, it does not require simultaneous operation of two cameras or an overlapping field of view. Compared to translation-based methods, it is extremely lightweight (2.7K vs. 26.1M parameters~\cite{perevozchikov2024}), supports various 
color transformations (e.g. 3$\times$3, root polynomials~\cite{finlayson2015}), and can be initialized with calibration measurements. Additionally, when used with an auxiliary illumination sensor, the model can be trained without paired data — meaning it does not require images of the same scene captured by both cameras.\\

\textbf{The contributions} of this work are as follows

\begin{enumerate}
 \item We introduce the NPM, a lightweight, physically-informed model that simulates the raw images of a color target captured by different cameras under a given illumination.
 It supports initialization from camera calibration data, offers compatibility with various color transformations, and can be trained on unpaired data.
 \item We show that illumination provided by an auxiliary sensor, such as a spectrometer or a multi-spectral sensor, allows the model to create a raw-to-raw mapping without requiring overlapping fields of views or simultaneous operation of both cameras.
 \item We demonstrate that our method outperforms recent state-of-the-art~(SotA) methods on public datasets such as NUS-8~\cite{cheng2014,afifi2021} and BeyondRGB~\cite{glatt2024}.
\end{enumerate}

\section{Related Work}
\label{sec:related_work}

\noindent\textbf{Image Formation}. The image \( I(x,c) \) at pixel \( x \) and color channel \( c \) is generated by integrating the product of three factors over the visible spectrum \( \omega \) as 

\begin{equation}
I(x,c) = \int_{\omega} S(c, \lambda) R(x, \lambda) L(\lambda) \, d\lambda
\label{eq:image_formation_continuous}
\end{equation}

where \( L(\lambda) \), the \textit{spectral power distribution} of the illumination; \( R(x, \lambda) \), the \textit{spectral reflectance} of the scene at pixel \( x \); and \( S(c, \lambda) \), the \textit{camera's spectral sensitivity} for color channel \(c\) 

Different cameras have distinct response functions, such as \( S^s(c,\lambda) \) and \( S^t(c,\lambda) \), due to variations in their optics, filters, and sensors. Manufacturers typically do not provide the exact spectral sensitivities of their cameras, and accurately measuring them requires costly equipment. While some studies~\cite{solomatov2023, tominaga2021} attempt to estimate these sensitivities using ColorChecker measurements, these methods often lack precision. 
Because of these differences in spectral sensitivities, the same object can appear different when captured by different cameras. To address these discrepancies, common approaches include calibration-based, translation-based, and photometric alignment methods.\\

\noindent\textbf{Calibration-based methods} chromatically align images acquired by multiple cameras relying on the results of offline camera calibration~\cite{ilie2005,joshi2005,sabater2017,wilburn2005,nguyen2014}. These methods adjust each camera's response using black and flat-field images and match their chromaticity via a transformation, typically a 3$\times$3 matrix calculated from a ColorChecker. In~\cite{nguyen2014}, the authors explore more complex transformations and emphasize the importance of illumination-dependent transforms. Based on this, they develop a raw-to-raw transformation method in an illumination-compensated color space. While calibration-based methods are efficient, computed offline, and therefore do not require images from both cameras, they  struggle under varying illumination conditions as noted in~\cite{nguyen2014}.

\noindent\textbf{Translation-based methods}~\cite{afifi2021,perevozchikov2024} treat the raw-to-raw transformation as an image translation task~\cite{zhao2020, zhu2017, park2020}, largely ignoring the physics of image formation and instead relying on deep learning. These methods are trained using either a mix of unpaired and limited paired data~\cite{afifi2021}, with latent feature matching, or entirely unpaired data~\cite{perevozchikov2024}, employing generative adversarial and cyclic losses. Their proposed training with unpaired data removes the costly requirement of capturing each scene with multiple cameras. However, despite their effectiveness, translation-based approaches are computationally demanding for edge devices and require large amounts of training data, partly because they are not physically informed. Also these approaches perform an implicit estimation of the spectral power distribution of the illuminant from RGB data, which is a inherently ill-posed problem.   

\noindent\textbf{Photometric alignment} approaches~\cite{liu2014,anirudth2018,song2022}, commonly used in image fusion and panoramic imaging, address raw-to-raw transformations by leveraging overlapping image regions to compute transformation parameters. These methods typically involve two main steps: image feature matching and transformation estimation. For image feature matching, these approaches rely either on block matching~\cite{liu2014} or on sparse image feature matching techniques~\cite{anirudth2018, song2022}. From the matched image features, these methods estimate transformations to maximize feature similarity across images, employing techniques such as gain adjustment, channel-wise mapping, mean/standard deviation alignment and histogram matching. These methods show strong results; however, they require overlapping and similar field of view and simultaneously operating cameras, and rely on a robust feature matching method.

\noindent\textbf{Color constancy} is a feature of the human visual system that helps maintain the perceived color of objects across different illumination conditions. To replicate this feature in a camera, color constancy methods first estimate the white point—the raw RGB values of an achromatic patch viewed under the given illumination. Using a Chromatic Adaptation Transform (CAT), 
these methods then adjust the colors to partially or fully account for the illumination, simulating human chromatic adaptation.

Traditional color constancy methods estimate the white point from image statistics~\cite{cheng2014, barnard2000, chakrabarti2011}, while modern approaches are often learning-based~\cite{barron2015, hu2017, yu2020}. In this work, we use the white point as a proxy for the illumination. We  estimate the white point using both the simple \textit{gray world} method~\cite{forsyth2002} and the state-of-the-art learning-based C5 method~\cite{afifi2021_1}, which adapts to a specific camera model using just a few example raw images.

\section{Method}
\label{sec:method}
Let us assume we have a \textit{source} and a \textit{target} camera with different spectral sensitivities, and, optionally, an auxiliary illumination sensor, such as a multi-spectral camera or a spectrometer. The source and target cameras may not share the same field of view. Given this setup, our objective is to find an illumination adaptive \textit{raw-to-raw} transformation \( \mathcal{F}_{\mathbf{L}}(\cdot) \), that transforms a raw source camera image \(\mathbf{I}^s \in \mathbb{R}^{3 \times W \times H} \) to resemble a hypothetical raw target camera image \(\mathbf{I}^t \in \mathbb{R}^{3 \times W \times H} \) that would have been acquired under given illumination if the field-of-view of the target camera matched the source camera:

\begin{equation}
\mathbf{I}^{s \rightarrow t} = \mathcal{F}_{\mathbf{L}}(\mathbf{I}^s) \quad \text{such that} \quad \mathbf{I}^{s \rightarrow t} \approx \mathbf{I}^t
\label{eq:problem_formulation}
\end{equation}

\begin{figure*}[h!]
\centering
\begin{minipage}{0.6\textwidth}
    \includegraphics[width=\textwidth]{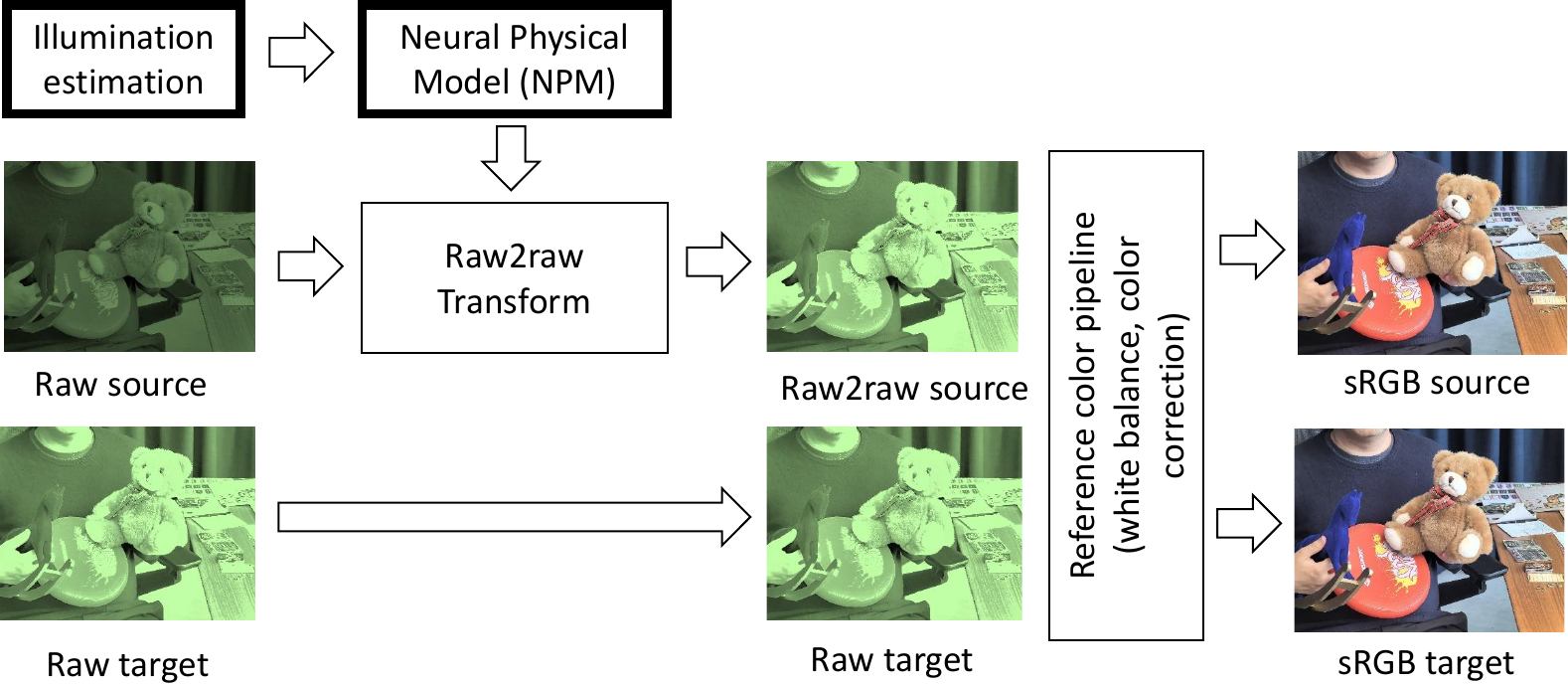}
\end{minipage}%
\hspace{0.02\textwidth}
\begin{minipage}{0.37\textwidth}
\caption{\textbf{System overview}. Our system consists of two main modules: Illumination Estimation and Neural Physical Model shown in bold. The Illumination Estimation module estimates the scene’s illumination, which the Neural Physical Model uses to compute an illumination adaptive raw-to-raw transformation. This transformation is applied to the source camera’s raw image, chromatically aligning it with the target camera’s raw image to enable a shared downstream color pipeline for both. }
\label{fig:system_overview}
\end{minipage}
\end{figure*}

\subsection{Network Architecture}

\noindent\textbf{System Overview}. To address the outlined problem, we propose a physically inspired network illustrated in Figure~\ref{fig:system_overview}, consisting of two main modules: Illumination Estimation and the Neural Physical Model. The Illumination Estimation module estimates information about the scene illumination either from a source image or an auxiliary  multi-spectral sensor and passes this information to the NPM. The NPM then computes an illumination adaptive raw-to-raw transformation based on this information. This transformation is applied to the raw source image, aligning it closely with the hypothetical target image. This alignment enables consistent downstream color processing, including white balance and color correction, for both images.\\

\begin{figure*}[h]
    \centering
    \begin{minipage}{0.6\textwidth}
        \includegraphics[width=\textwidth] {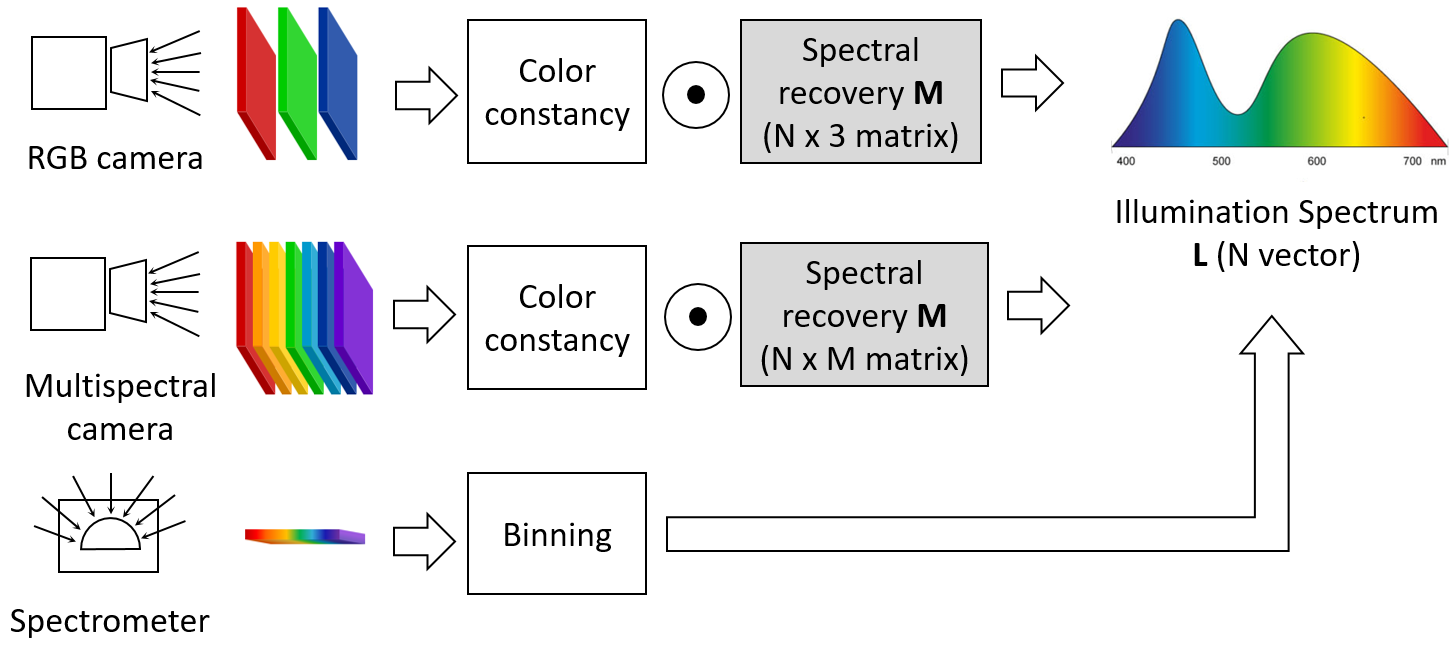} 
    \end{minipage}%
    \hspace{0.02\textwidth}
    \begin{minipage}{0.37\textwidth}
        \caption{The \textbf{Illumination Estimation} module estimates the spectral power distribution~(SPD) of the scene illumination using either an RGB source camera, or an auxiliary multi-spectral camera or spectrometer. For RGB and multi-spectral cameras, a color constancy method is first applied to the images to determine the illumination’s white point, which is then converted to an SPD using a learnable spectral recovery matrix shown in gray. For spectrometer data, the measured SPD is simply binned into finer spectral bins.}
        \label{fig:illumination_estimation}
    \end{minipage}
\end{figure*}

\noindent The \textbf{Illumination Estimation} module, illustrated in Figure~\ref{fig:illumination_estimation}, estimates the spectral power distribution  of the illumination using either the raw source camera image, or an auxiliary multi-spectral camera, or a spectrometer. We note that such auxiliary sensors have recently become available in consumer devices~\cite{koskinen2024}. When using an image from the source RGB or multi-spectral camera, we first apply a color constancy method (C5~\cite{afifi2021_1} or simple gray world) to estimate the illumination’s white point, which we then convert to an SPD, denoted as \( \mathbf{L} \in \mathbb{R}^{N} \). We express this procedure as follows:

\begin{equation}
\mathbf{L} = \mathbf{M} \cdot \operatorname{CC}(\mathbf{I}), 
\label{eq:illumination_estimation}
\end{equation}

\noindent where \( \mathbf{M} \in \mathbb{R}^{N \times M}  \) is a learnable \textit{spectral recovery matrix}, \(\operatorname{CC}(\cdot)\) denotes the \textit{color constancy method}, and \(\mathbf{I} \in \mathbb{R}^{M \times W \times H} \) is the image used for illumination estimation. When using spectrometer data, we compute the SPD by interpolating the measured spectral values into finer bins.\\

\begin{figure*}[h]

\centering
\begin{minipage}{0.64\textwidth}
 \includegraphics[width=\textwidth]{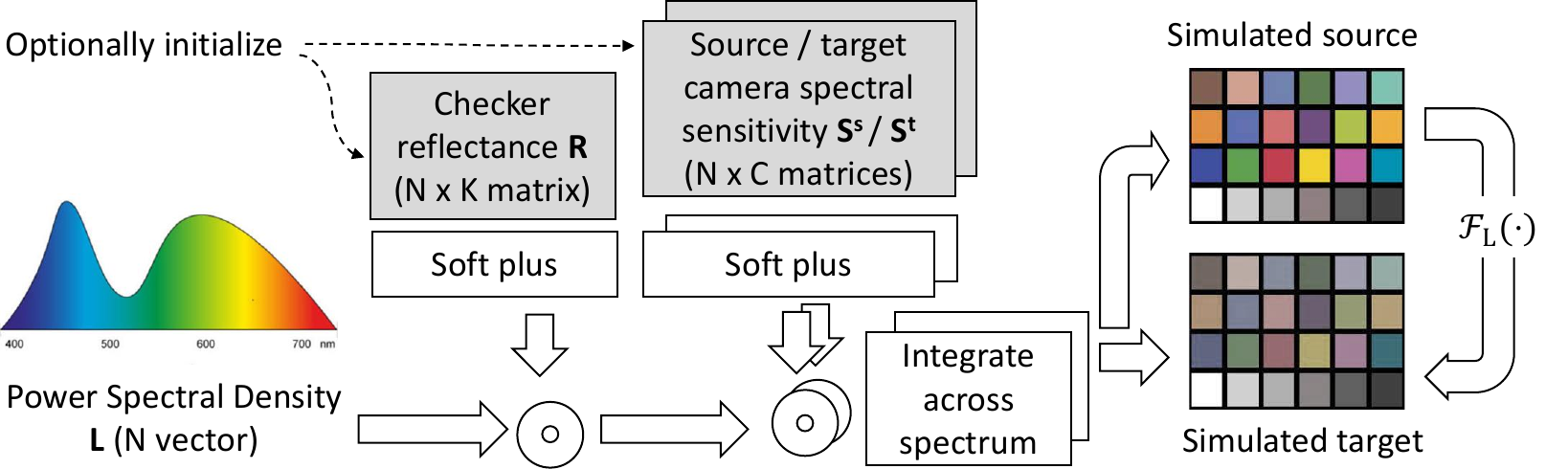}
\end{minipage}%
\hspace{0.02\textwidth}
\begin{minipage}{0.33\textwidth}
    \caption{\textbf{Neural Physical Model~(NPM)} takes the spectral power distribution of the scene illumination as input and simulates raw images of a ColorChecker under this illumination, as captured by the source and target cameras, using a physical image formation model. These simulated checker images we  then use to estimate the illumination-specific raw-to-raw transformation between the source and target cameras. The learnable parameters of the NPM, shown in gray, can optionally be initialized with camera calibration data. }
\label{fig:neural_physical}
\end{minipage}
\end{figure*}

\noindent The \textbf{Neural Physical Model}, shown in Figure~\ref{fig:neural_physical}, takes the spectral power distribution as input and generates raw images of a color target, simulating how they would be captured by the source and target cameras under the specified illumination. This process follows the physical image formation model described in Section~\ref{sec:related_work} and Equation~\ref{eq:image_formation_continuous}. In matrix form, this can be represented as:

\begin{equation}
\mathbf{\hat{I}}_x = \mathbf{S} \cdot \mathbf{R}_x \cdot \mathbf{L},
\label{eq:image_formation_matrix}
\end{equation}

\noindent where \( \mathbf{\hat{I}}_x \in \mathbb{R}^{3} \) is the vector of simulated raw image values for the ColorChecker patch \( x \), \( \mathbf{S} \in \mathbb{R}^{3 \times N} \) is the learnable \textit{camera spectral sensitivity} matrix, and \( \mathbf{R}_x \in \mathbb{R}^{N \times N} \) is a learnable diagonal matrix representing the \textit{spectral reflectance} of patch \( x \). These matrices can be optionally initialized from calibration data available to the camera manufacturer. Note, that while we use the \textit{Macbeth} ColorChecker~\footnote{Macbeth ColorChecker, Munsell Color Science Laboratory, 1976. Available from X-Rite, Inc. at \url{https://www.xrite.com}}, our method is compatible with other color targets as well.

Next, we estimate the illumination-adaptive raw-to-raw transformation from the simulated ColorChecker images using a least squares estimator, normalizing both images by the green channel value of the \textit{Neutral 8} patch to minimize dependency on illumination intensity and account for the prevalent over-saturation of the \textit{White} patch in measurements. 

In this work, we adopt a simple and stable illumination-adaptive linear transformation \(\mathcal{F}_{\mathbf{L}} \in \mathbb{R}^{3 \times 3}\), which has been shown to perform well compared to more complex transformations~\cite{nguyen2014}. However, the proposed method is not restricted to linear transformations and can be readily adapted for use with other transformations such as described in~\cite{finlayson2015, finlayson2016}.

Note that if the illumination is estimated using an auxiliary sensor, the NPM can be trained separately for each camera. This enables \textit{unpaired training}, as in~\cite{afifi2021,perevozchikov2024}, which does not require a training dataset where each scene is captured by two cameras. This approach is practical because it allows new camera models to be added without re-capturing the entire training dataset, which is especially useful when updating or expanding a product line with a new camera. 



\subsection{Training Losses}
The proposed method supports both \textit{paired} and \textit{unpaired} training modes. In paired mode, it requires scenes with a ColorChecker captured under different illuminations by both source and target cameras. In unpaired mode, each scene can be captured by only one camera, enabling easy extension of existing datasets with new camera modules without recapturing data for other modules. Note that our training procedure uses only checker images, allowing the entire training dataset to be collected in a laboratory with a tunable light source in a very short time. 

The model is trained with two goals: Firstly, given a measured illumination, to generate raw images of ColorCheckers as they would have been captured by source and target cameras under that illumination. This goal is implemented in our simulation loss that can be used in paired and unpaired training mode. 
Secondly, given a measured illumination, to generate a transformation that can be applied to the source camera raw ColorChecker image, so that in XYZ space it matches the image from the target camera.
Note, that the transformation is estimated purely from generated raw checker images. This goal is implemented in our Matching Loss, that can be used only in the paired mode. Further details on these losses are provided below.\\

\noindent The \textbf{Simulation Loss} encourages the model to accurately simulate raw image values for checkers for both the source and target cameras under a given illumination. Since the simulation loss is defined separately for each camera, it can be used in paired and unpaired training, where training scenes are not captured simultaneously by both cameras. The simulation loss is computed as

\begin{equation}
L^i_{\text{sim}} = \frac{1}{K} \sum_{x=0}^{K-1} \delta(\mathbf{I}^i_x, \mathbf{\hat{I}}^i_x) + w \frac{1}{K} \sum_{x=0}^{K-1} |\mathbf{I}^i_x - \mathbf{\hat{I}}^i_x|
\label{eq:simulation_loss}
\end{equation}

\noindent where \(\mathbf{\hat{I}}^i_x\) and \(\mathbf{I}^i_x\) are vectors of the simulated and measured raw image values for ColorChecker patch \(x\) and camera \(i \in \{r, s\}\); \(\delta(\cdot, \cdot)\) is the angular error function, which computes the angle between two vectors; \(K\) is the total number of ColorChecker; and \(w\) is the weight of the \(L_1\) loss. We use a combination of \(L_1\) and angular loss to encourage the network to preserve the relative intensity between checkers. \\

\noindent The \textbf{Matching Loss} encourages the model to produce a raw-to-raw transformation that minimizes the difference between ColorChecker values from the transformed source and the target cameras. 
This loss is computed in the human perception-related XYZ color space. 
To convert raw images to XYZ, we use an interpolated color correction matrix derived from two predefined matrices (illuminants D65 and A) found in the metadata of the target camera, as described in~\cite{karaimer2018}. 
Note, that this loss relies on ColorChecker images captured by source and target cameras under the same scene illumination, therefore it can not be used for unpaired training. Mathematically, the matching loss is computed as

\begin{align}
\begin{split}\label{eq:matching_loss}
L_{\text{match}} = &\frac{1}{K} \sum_{x=0}^{K-1} \delta( \mathbf{C} \cdot \mathcal{F}_{\mathbf{L}}(\mathbf{I}^s_x), \mathbf{C} \cdot \mathbf{I}^t_x)\\
&+ w \frac{1}{K} \sum_{x=0}^{K-1} |\mathbf{C} \cdot \mathcal{F}_{\mathbf{L}}(\mathbf{I}^s_x) - \mathbf{C} \cdot \mathbf{I}^t_x|, 
\end{split}
\end{align}

\noindent where \(\mathbf{I}^s_x\) and \(\mathbf{I}^t_x\) are the raw image vector values of ColorChecker patch \(x\) as captured by the source and the target camera respectively, \(\mathbf{C}\) is the color correction matrix transforming raw values to the XYZ color space.

\section{Experiments}
\label{sec:experiments}

\noindent\textbf{Evaluation Metrics}. To evaluate the performance of our method, we apply the raw-to-raw transformation to the Macbeth ColorChecker values captured by the source camera. We then map both the transformed source camera checker values and the target camera checker values to XYZ color-space using the same approach as for our Matching Loss. Finally, we transform both checkers to Lab space, using the measured target checker \textit{Neutral 8} patch as illumination estimation and compare them using the \textit{CIEDE2000}~\cite{sharma2005}, which we denote as \( \Delta E \), to account for perceptual differences in color. \\

\noindent\textbf{Datasets}. We use two public datasets: NUS-8~\cite{cheng2014} and BeyondRGB~\cite{glatt2024}, and one internal \textit{Huawei Pura 70} dataset, collected with the Huawei Pura 70 Ultra. We do not use the Raw-to-Raw dataset~\cite{afifi2021}, as it lacks the original checker images. 

The \textit{NUS-8}~\cite{cheng2014} dataset, initially developed for evaluating color constancy methods, contains around 250 indoor and outdoor scenes, each captured using eight different cameras with a Macbeth ColorChecker positioned in the corner of each frame. Following the protocol in~\cite{afifi2021}, we restrict our use to the Canon EOS600D and Nikon D5200 cameras and manually identify corresponding scenes between the two. 
We designate the Canon EOS600D as the target camera and randomly select a subset of 192, 35, and 35 images for training, validation, and testing, respectively, as the referenced original experimental protocol~\cite{afifi2021, perevozchikov2024} does not specify a dataset split.   

The \textit{Beyond RGB}~\cite{glatt2024} dataset consists of approximately 1800 scenes captured under both laboratory and real-world conditions, using Samsung Galaxy S21 Plus and Oppo Find X5 Pro smartphones, along with a 16-filter multi-spectral camera and a 36-channel spectrometer illumination measurement. For each scene, a separate image containing a Macbeth ColorChecker is captured by each device. A key advantage of this dataset is its inclusion of smartphone images with substantial color variations, auxiliary illumination sensor data, and scenes captured under diverse lighting conditions. In our experiments, we designate the Oppo Find X5 Pro as the target camera and use the training, validation, and testing split provided in~\cite{glatt2024}.

Finally, we collected the \textit{Huawei Pura 70} dataset to demonstrate the practical applications of our method, such as achieving a color-consistent multi-camera zoom. The phone itself, a Huawei Pura 70 Ultra, has three cameras with distinct color and field-of-view variations, as well as an 8-channel single-pixel multi-spectral sensor. For each scene, we saved raw images from all cameras along with the corresponding multi-spectral data. For training, we captured images of a Macbeth ColorChecker under various illuminations from a tunable light source, sampling the illumination uniformly in the \(xy\) color space and also along the Planckian locus. For testing we collect additional real-world scenes, each paired with an image containing a ColorChecker. In total, we collect 402 images under 134 different illuminations. For this dataset, we designate the main camera as our target camera.\\

\noindent\textbf{Experiment setup}. All experiments were conducted using the PyTorch deep learning framework~\cite{paszke2019} and the Colour Science library~\cite{colour2022}, which provides color-related functions. The models are trained using the Adam optimizer with default parameters 
and a learning rate of 0.01, which decreases on plateaus. Training typically requires around 100 epochs with a batch size of 4. 
We initialize the models with measured Macbeth ColorChecker reflectances~\footnote{Spectral reflectances for the Macbeth ColorChecker are available at \url{https://babelcolor.com}}, but except for the Huawei Pura 70 dataset, we do not initialize camera spectral sensitivities, as they are typically unavailable.

\subsection{Ablations}
\label{sec:ablations}

In this section we investigate the importance of different components of our method. All ablation results are summarized in Table~\ref{tab:ablations}.

\noindent\textbf{Illumination estimation}. In this section, we investigate the role of illumination in raw-to-raw mapping. We examine the importance of illumination estimation methods and the spectral resolution of the illumination sensor. We perform this ablation on the BeyondRGB dataset, as it contains auxiliary illumination sensors, as well as scenes with diverse illumination. 
As shown in the Table~\ref{tab:ablations}, the illumination-agnostic raw-to-raw transformation performs the worst. The 3$\times$3 transform computed in white-balanced space as in Raw-to-Raw~\cite{nguyen2014} already performs slightly better, but it is still worse than our proposed method with a 3-channel image for illumination estimation. This confirms that accurate illumination information is essential for effective raw-to-raw mapping between cameras. Notably, replacing the naive grey world color constancy method (angular error of 6.72° towards the \textit{Neutral 8} patch) with the state-of-the-art C5~\cite{afifi2021_1} method (angular error of 3.92°) improves the results by a larger margin in comparison, as the illumination estimation provided is more accurate. The largest performance boost, however, is achieved with sensors offering higher spectral resolution, such as a 16-channel multi-spectral camera or, ideally, a spectrometer with 36 channels.

\noindent\textbf{Paired Training}. As described in Section~\ref{sec:method}, the proposed method, when used with an auxiliary illumination sensor, supports both unpaired and paired training modes. The main advantage of the unpaired mode is that it does not require a dataset with aligned images from both cameras for each scene, making it practical to add new camera models without re-capturing the entire training dataset. In contrast, paired training enables the use of an additional matching loss, allowing the model to directly minimize the perceptual difference between the transformed source camera image and the target camera image. As shown in Table~\ref{tab:ablations}, the proposed method, which uses a multi-spectral sensor with the gray-world method for illumination estimation, outperforms the illumination-agnostic method in both paired and unpaired training modes. 

\noindent\textbf{Physically informed initialization}. Since our model is physically-informed, we can initialize it with values measured during camera calibration, such as checker reflectances, camera spectral sensitivities and a spectral recovery matrix. This initialization should provide a strong starting point, with fine-tuning further enhancing performance by correcting for measurement imperfections and co-adapting parameters to optimize the results. 
Unfortunately, measuring camera spectral sensitivities typically requires expensive equipment and is usually only performed by camera manufacturers, who rarely make this data publicly available. As a result, these measurements are often unavailable for cameras in public datasets.
To address this limitation, we perform this ablation on our newly collected Huawei Pura 70 dataset, for which we measured the camera sensitivities ourselves, using the 8-channel single-pixel sensor as illumination estimation. As shown in Table~\ref{tab:ablations}, initializing the model with calibration parameters significantly outperforms the illumination-agnostic baseline. Notably, additional training further enhances performance, resulting in a model that is more stable under extreme illumination conditions. Furthermore, initializing the model from measured values leads to faster convergence and improved results compared to training from random initialization.

\begin{table}[htb]
\centering
\begin{tabularx}{\columnwidth}{ lR }
\hline
\textbf{Methods} & \( \mathbf{\Delta E \downarrow } \)  \ \\
\hline
\multicolumn{2}{c}{Illumination estimation~(BeyondRgb - Oppo to Samsung)} \\
\hline
Illumination agnostic (3$\times$3) & 3.48 \\
Raw-to-Raw~\cite{nguyen2014} & 3.41 \\
Grey World, 3 channel sensor  & 3.37 \\
C5~\cite{afifi2021_1}, 3 channel sensor & 3.29 \\
Grey World, 16 channel sensor & 3.11 \\
\textbf{Spectrometer, 36 channels sensor} & \textbf{2.94} \\ \hline
\multicolumn{2}{c}{Paired training~(BeyondRgb - Oppo to Samsung)} \\
\hline
Illumination agnostic (3$\times$3) & 3.48 \\
Unpaired training & 3.21 \\ 
\textbf{Paired training} & \textbf{3.11} \\
\hline
\multicolumn{2}{c}{Physically informed initialization~(P70 - Wide to Main)} \\
\hline
Illumination agnostic (3$\times$3) & 3.10 \\
Initialized, before training  & 2.56 \\
Non-initialized, after training & 2.49 \\
\textbf{Initialized, after training} & \textbf{2.30} \\

\hline
\end{tabularx}
\caption{Ablation study summary. Each ablation specifies the dataset used. We include an illumination-agnostic method as a baseline, which uses a single linear matrix optimized on the entire training set. The best-performing method is highlighted in bold.}
\label{tab:ablations}
\end{table}

\subsection{Benchmarking}
\label{sec:benchmarking}

To compare our method to other existing methods we use the NUS-8 dataset, which has current results performed by using a Semi-Supervised method~\cite{afifi2021}, which is trained on a matching set of 10 images, as well as Rawformer~\cite{perevozchikov2024}, trained in a fully unsupervised manner. Both of them have their own advantages, with either using a large model such as Rawformer (26.1M parameters) for unpaired training, or using a slightly smaller Encoder-Decoder setup with anchor images. 
As there is no auxiliary illumination sensor for this dataset, we defer to the grey world illumination estimation from our source Nikon image, to generate both checkers in an unpaired manner. 
We direct to our ablations on the BeyondRGB dataset to draw parallels on the benefits of using additional spectral information for this dataset.
As the reported methods are not directly reproducible, we refer to their reported results and tried our best to ensure an honest comparison.

\begin{figure}[tb]
  \centering
  \begin{subfigure}{0.49\linewidth}
    \centering
    \includegraphics[width=\linewidth]{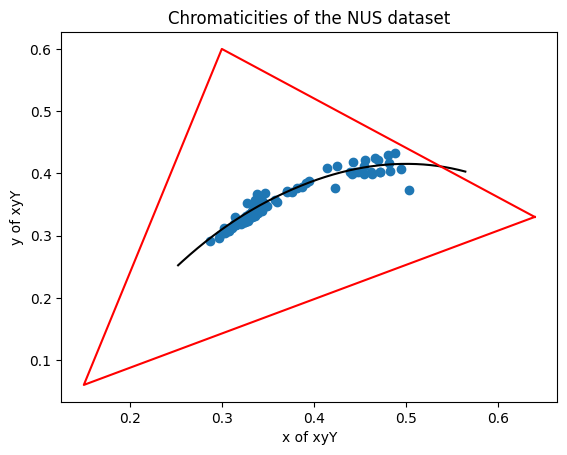}
    \caption{NUS-8 training set.}
  \end{subfigure}
  \begin{subfigure}{0.49\linewidth}
    \centering
    \includegraphics[width=\linewidth]{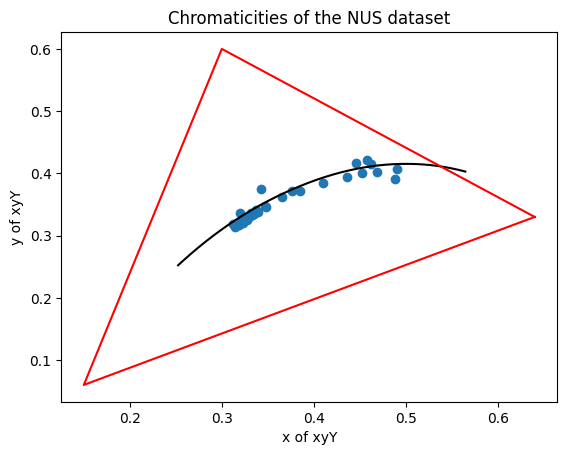}
    \caption{NUS-8 validation set.}
  \end{subfigure}
  \begin{subfigure}{0.49\linewidth}
    \centering
    \includegraphics[width=\linewidth]{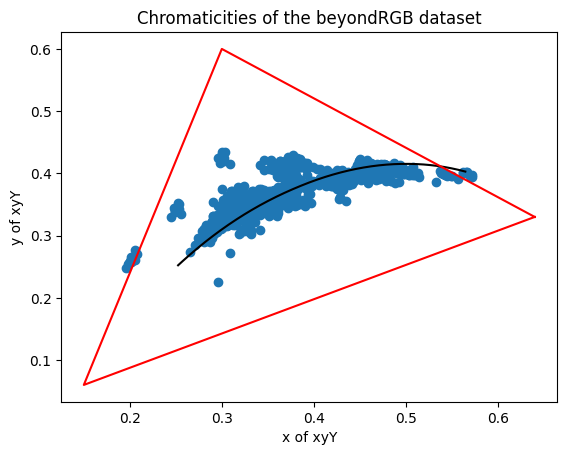}
    \caption{BeyondRGB training set.}
  \end{subfigure}
  \begin{subfigure}{0.49\linewidth}
    \centering
    \includegraphics[width=\linewidth]{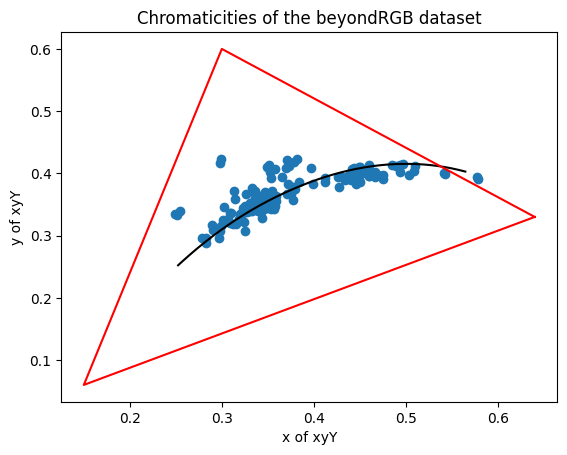}
    \caption{BeyondRGB validation set.}
  \end{subfigure}
  \caption{xyY distribution of the NUS-8 and BeyondRGB dataset illumination for the given training and validation sets. The red triangle represents the sRGB space and the Plackian curve is denoted in black. xyY chromaticities are estimated with the given XYZ transformation matrices.}
  \label{fig:distribution}
\end{figure}

\noindent\textbf{Dataset Chromaticity.} 
In regards to raw-to-raw benchmarks, dataset diversity is often an under-addressed issue, which should be considered, as it is one of the main factors affecting translation accuracy. 
To this regard we explicitly analyze the diversity of our two selected datasets in Figure~\ref{fig:distribution}. 
Beyond the relative sparsity of the data in NUS-8 compared to BeyondRGB, we note that large sections of the chromaticity-space are not considered, making the raw-to-raw mapping a relatively trivial task for simple networks, as this is not the originally intended purpose of this dataset. We therefore propose moving future raw-to-raw comparisons to more diverse and illumination-rich datasets such as BeyondRGB, where the benefits of different methods become clearer.

\begin{table}[ht]
\centering
\begin{tabularx}{\columnwidth}{ lRr }
\hline
\textbf{Methods (NUS - Nikon to Canon)} & \textbf{$\Delta$E} & \textbf{Model Size} \\  
\hline
Semi-Supervised~\cite{afifi2021} & 5.95 & Unreported \\
Rawformer~\cite{perevozchikov2024} & 2.53 & 26.1M \\
NPM with Grey World~(ours) & 1.76 & 2.7K \\
\hline

\end{tabularx}
\caption{Comparative results on the NUS-8 dataset. For the Semi-supervised and Rawformer method, we use results reported in~\cite{afifi2021,perevozchikov2024}.} 
\label{tab:nus}
\end{table}

\section{Discussion and Future Work}

In this work we have shown that using a small physically-informed network, we are able to boost raw-to-raw mapping accuracy by a significant margin. The network structure also allows for selective training, fixing measurements that we are certain of, and training other parameters based on availability and pairs of images.
Beyond using larger models and more complex transformations (other than a 3$\times$3 matrix), these illumination-dependent networks could also be used for pixel-local color-transformations, based on local white-balance maps in multi-illumination cases.
Finally our model can also be made to be content-adaptive by generating patches based on the content of the image, instead of being restricted to the Macbeth ColorChecker.

{\small
\bibliographystyle{ieee_fullname}
\bibliography{egbib}
}

\section{Supplementary Material}

\subsection{Examples}
For visual qualitative evaluation, we present eight test examples comparing images captured directly from the latest available flagship phones and compare them with the output of our physically inspired network. The outputs from the wide and main camera are cropped, as in Figure \textbf{1} of our paper, to more easily compare color differences.

The phone results are the output of a full Image Signal Processing (ISP) pipeline, which typically includes advanced features such as multi-exposure fusion, denoising, 3D look-up tables, and photometric alignment from multiple active cameras.
In contrast, our algorithm operates within a simplified ISP pipeline. This pipeline includes demosaicing, black level subtraction, normalization, raw-to-raw transformation (mapping from wide to main or tele to main) with our network, and automatic white balance (AWB) with color correction in the main camera's raw color space. Notably, we exclude chromatic adaptation transforms (CAT) for simplicity, as the primary focus is on the relative differences between images rather than absolute color fidelity.

The focus of this setup are the differences between columns, which illustrate the effectiveness of our raw-to-raw transformations, while differences between rows reflect the inherent disparities in raw-to-RGB transformations across cameras.

We observe significant discrepancies, particularly between the telephoto and main cameras in current phones. Examples \textbf{4} and \textbf{5} further illustrate scenarios with multiple distinct illuminants, presenting additional challenges for accurate processing.
In these cases, our Neural Physical Model (NPM) also exhibits visible color differences. We attribute this to the multispectral grey-world AWB assumption, which relies on the Pura70 multispectral singular pixel as the illumination estimation input for the NPM. This grey-world assumption, while generally effective, may not accurately represent the overall illumination conditions in scenes with complex or mixed lighting environments.

\begin{table*}[ht]
\centering
\caption*{Example 1}
\vspace{-8pt}
\begin{tblr}{
  colspec={m{0.5cm}@{}X[c]@{}X[c]@{}X[c]@{}}, 
  width = \proportion\linewidth,
  rowspec={},          
  rowsep = 2pt,
  stretch = 0,
}
  \rotatebox{90}{\centering~~~~~~~iPhone 15 Pro Max} & 
  \includegraphics[width=\linewidth]{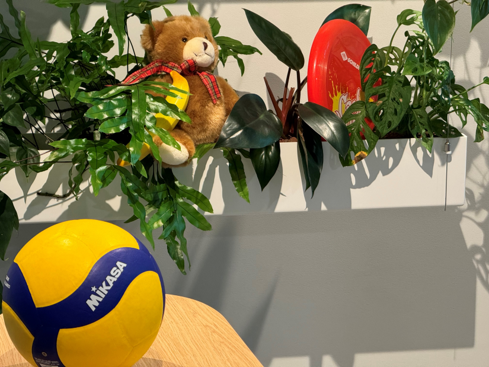} &
  \includegraphics[width=\linewidth]{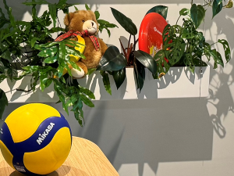} &
  \includegraphics[width=\linewidth]{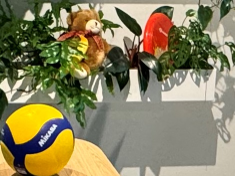} \\ \hline
  \rotatebox{90}{\centering~~~~~~~~~~~~~Pura70 Ultra} & 
  \includegraphics[width=\linewidth]{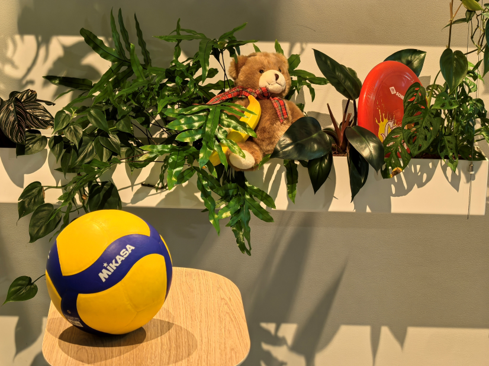} &
  \includegraphics[width=\linewidth]{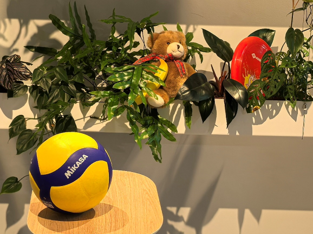} &
  \includegraphics[width=\linewidth]{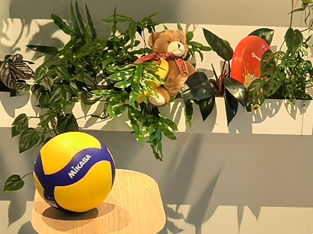} \\ \hline
  \rotatebox{90}{\centering~~~~~~~~~~~~~ NPM (Ours)} & 
  \includegraphics[width=\linewidth]{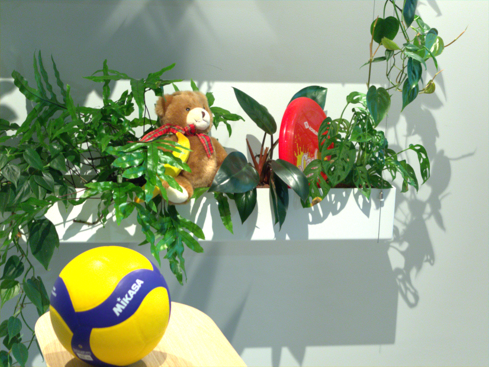} &
  \includegraphics[width=\linewidth]{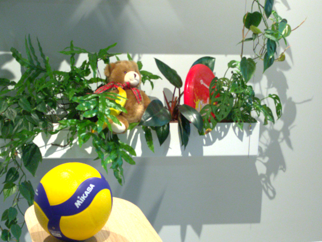} &
  \includegraphics[width=\linewidth]{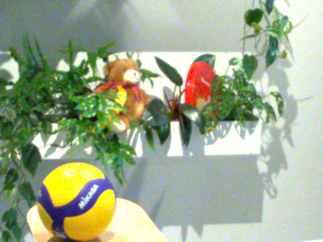} \\
  & Tele & Main & Wide \\
\end{tblr}
\caption*{
\generictext
In this example we can see a distinct shift of the iPhone tele camera table color compared to main and wide. We can also see a lesser shift between main and wide cameras in both iPhone and Pura70 phones.
}
\label{tab:qualitative_comparisons_2}
\end{table*}

\begin{table*}[ht]
\centering
\caption*{Example 2}
\vspace{-8pt}
\begin{tblr}{
  colspec={m{0.5cm}@{}X[c]@{}X[c]@{}X[c]@{}}, 
  width = \proportion\linewidth,
  rowspec={},          
  rowsep = 2pt,
  stretch = 0,
}
  \rotatebox{90}{\centering~~~~~~~iPhone 15 Pro Max} & 
  \includegraphics[width=\linewidth]{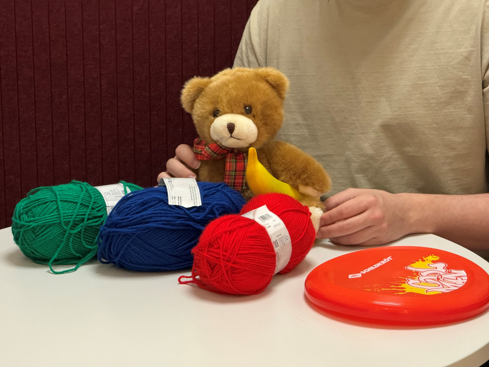} &
  \includegraphics[width=\linewidth]{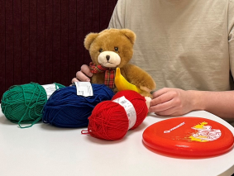} &
  \includegraphics[width=\linewidth]{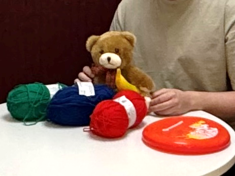} \\ \hline
  \rotatebox{90}{\centering~~~~~~~~~~~~~Pura70 Ultra} & 
  \includegraphics[width=\linewidth]{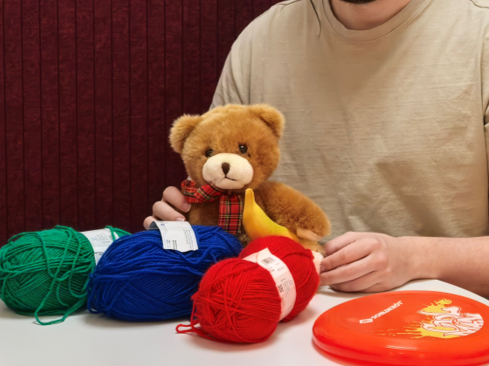} &
  \includegraphics[width=\linewidth]{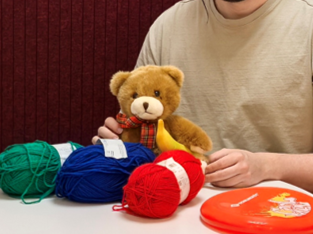} &
  \includegraphics[width=\linewidth]{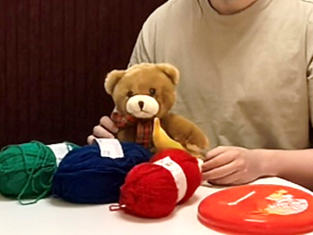} \\ \hline
  \rotatebox{90}{\centering~~~~~~~~~~~~~ NPM (Ours)} & 
  \includegraphics[width=\linewidth]{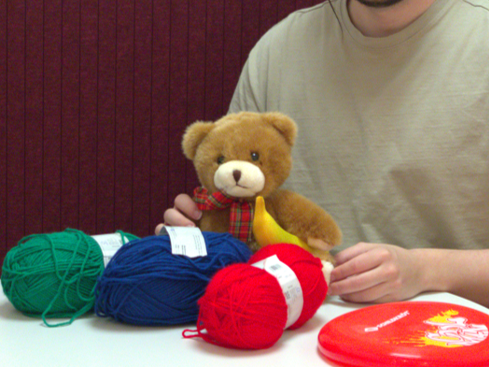} &
  \includegraphics[width=\linewidth]{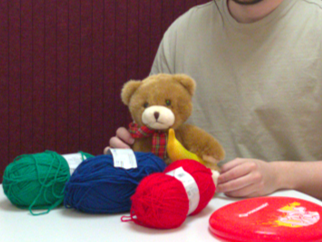} &
  \includegraphics[width=\linewidth]{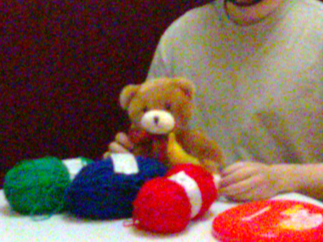} \\
  & Tele & Main & Wide \\
\end{tblr}
\caption*{
\generictext
In this example we can see a distinct shift of the iPhone tele camera table color to the main camera, and a lesser shift from main to wide, as seen on the T-shirt. For our algorithm, the comparison between main and wide is a bit harder to compare, as the single exposure wide-camera shot is very noisy (we do not use a AI-denoising algorithm).
}
\label{tab:qualitative_comparisons_3}
\end{table*}

\begin{table*}[ht]
\centering
\caption*{Example 3}
\vspace{-8pt}
\begin{tblr}{
  colspec={m{0.5cm}@{}X[c]@{}X[c]@{}X[c]@{}}, 
  width = \proportion\linewidth,
  rowspec={},          
  rowsep = 2pt,
  stretch = 0,
}
  \rotatebox{90}{\centering~~~~~~~iPhone 15 Pro Max} & 
  \includegraphics[width=\linewidth]{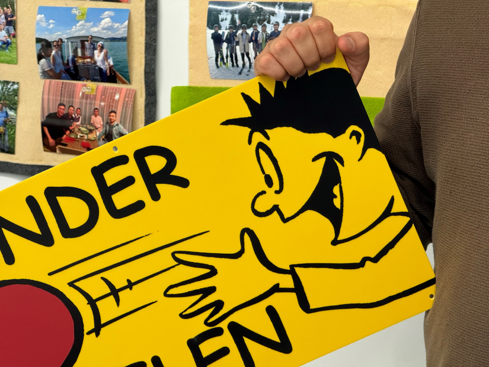} &
  \includegraphics[width=\linewidth]{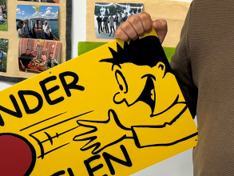} &
  \includegraphics[width=\linewidth]{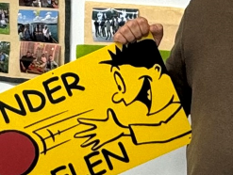} \\ \hline
  \rotatebox{90}{\centering~~~~~~~~~~~~~Pura70 Ultra} & 
  \includegraphics[width=\linewidth]{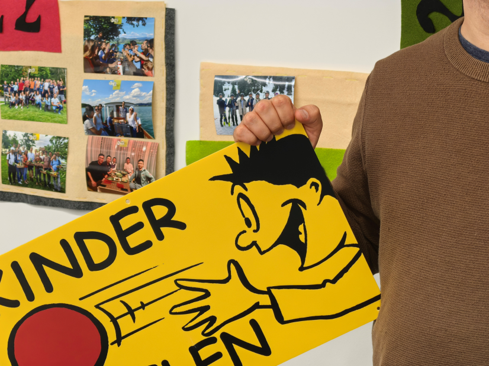} &
  \includegraphics[width=\linewidth]{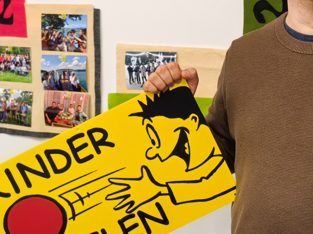} &
  \includegraphics[width=\linewidth]{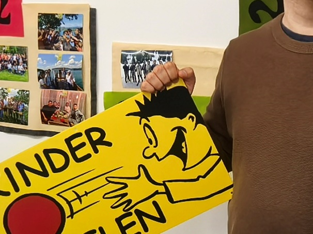} \\ \hline
  \rotatebox{90}{\centering~~~~~~~~~~~~~ NPM (Ours)} & 
  \includegraphics[width=\linewidth]{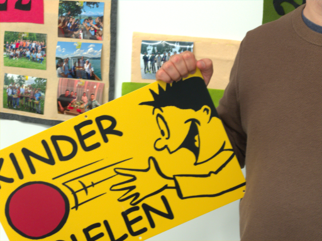} &
  \includegraphics[width=\linewidth]{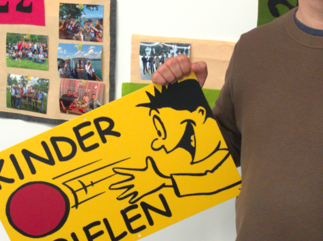} &
  \includegraphics[width=\linewidth]{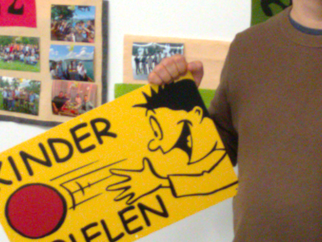} \\
  & Tele & Main & Wide \\
\end{tblr}
\caption*{
\generictext
In this example, which contains a simple illumination scene, all phones and our algorithm perform relatively well, with only a slight noticeable discoloration on the white background and hand in the Pura70 and, to an even lesser degree, on the iPhone.
}
\label{tab:qualitative_comparisons_4}
\end{table*}

\begin{table*}[ht]
\centering
\caption*{Example 4}
\vspace{-8pt}
\begin{tblr}{
  colspec={m{0.5cm}@{}X[c]@{}X[c]@{}X[c]@{}}, 
  width = \proportion\linewidth,
  rowspec={},          
  rowsep = 2pt,
  stretch = 0,
}
  \rotatebox{90}{\centering~~~~~~~iPhone 15 Pro Max} & 
  \includegraphics[width=\linewidth]{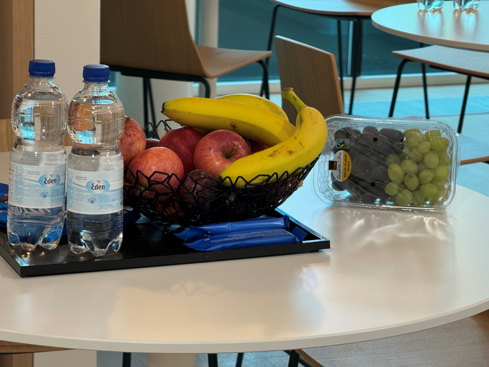} &
  \includegraphics[width=\linewidth]{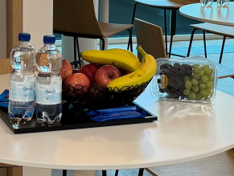} &
  \includegraphics[width=\linewidth]{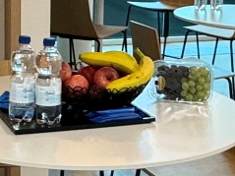} \\ \hline
  \rotatebox{90}{\centering~~~~~~~~~~~~~Pura70 Ultra} & 
  \includegraphics[width=\linewidth]{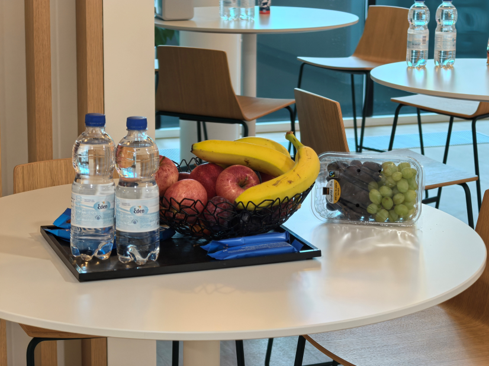} &
  \includegraphics[width=\linewidth]{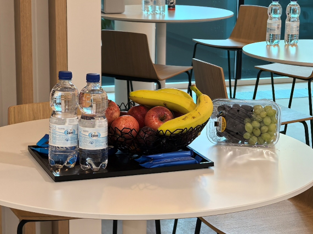} &
  \includegraphics[width=\linewidth]{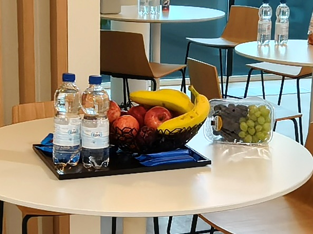} \\ \hline
  \rotatebox{90}{\centering~~~~~~~~~~~~~ NPM (Ours)} & 
  \includegraphics[width=\linewidth]{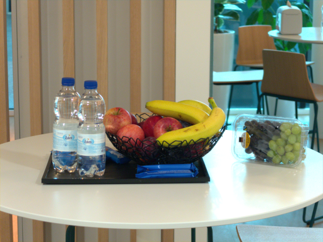} &
  \includegraphics[width=\linewidth]{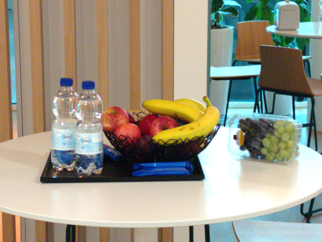} &
  \includegraphics[width=\linewidth]{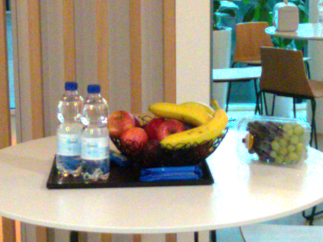} \\
  & Tele & Main & Wide \\
\end{tblr}
\caption*{
\generictext
In this multi-illumination example we can see all phones and algorithms failing, specifically our algorithm. As we use the multi-spectral pixel from the Pura70 as illumination estimation (grey-world assumption), we surmise that the estimated average illuminant taken from the whole scene, does not accurately represent the dominant illuminant in the tele-camera photo, which predominantly seems to be the reflection from the outside illuminant.
}
\label{tab:qualitative_comparisons_5}
\end{table*}

\begin{table*}[ht]
\centering
\caption*{Example 5}
\vspace{-8pt}
\begin{tblr}{
  colspec={m{0.5cm}@{}X[c]@{}X[c]@{}X[c]@{}}, 
  width = \proportion\linewidth,
  rowspec={},          
  rowsep = 2pt,
  stretch = 0,
}
  \rotatebox{90}{\centering~~~~~~~iPhone 15 Pro Max} & 
  \includegraphics[width=\linewidth]{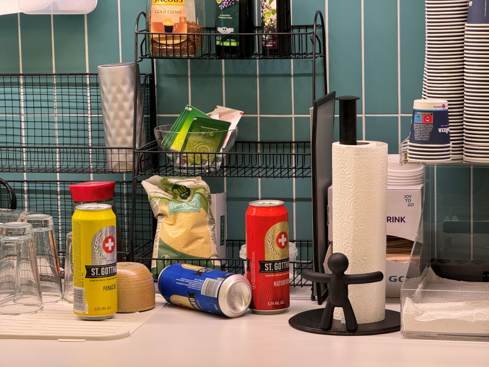} &
  \includegraphics[width=\linewidth]{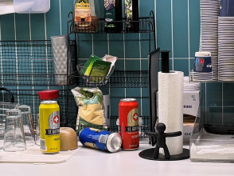} &
  \includegraphics[width=\linewidth]{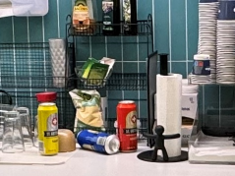} \\ \hline
  \rotatebox{90}{\centering~~~~~~~~~~~~~Pura70 Ultra} & 
  \includegraphics[width=\linewidth]{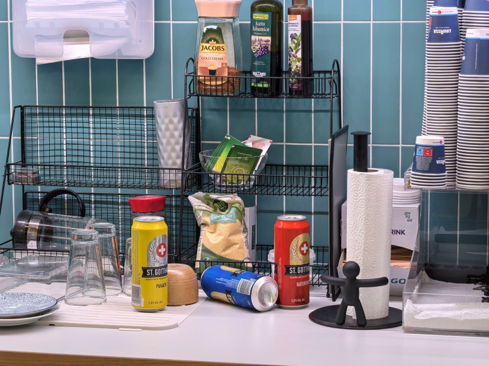} &
  \includegraphics[width=\linewidth]{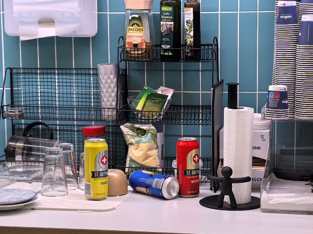} &
  \includegraphics[width=\linewidth]{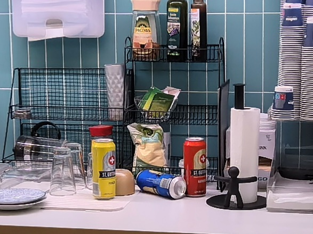} \\ \hline
  \rotatebox{90}{\centering~~~~~~~~~~~~~ NPM (Ours)} &  
  \includegraphics[width=\linewidth]{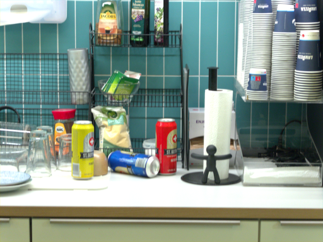} &
  \includegraphics[width=\linewidth]{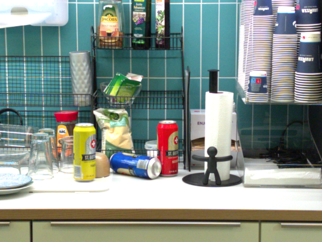} &
  \includegraphics[width=\linewidth]{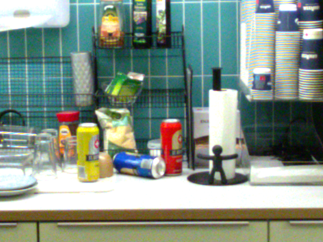} \\
  & Tele & Main & Wide \\
\end{tblr}
\caption*{
\generictext
In this example we see a dual illumination scene, where the global illumination does not fit the tele camera scene. In this example, while there are slight discrepancies in the table color for both the Pura70 and our algorithm, the iPhone seems to fail completely for the tele camera.
}
\label{tab:qualitative_comparisons_6}
\end{table*}

\begin{table*}[ht]
\centering
\caption*{Example 6}
\vspace{-8pt}
\begin{tblr}{
  colspec={m{0.5cm}@{}X[c]@{}X[c]@{}X[c]@{}}, 
  width = \proportion\linewidth,
  rowspec={},          
  rowsep = 2pt,
  stretch = 0,
}
  \rotatebox{90}{\centering~~~~~~~iPhone 15 Pro Max} & 
  \includegraphics[width=\linewidth]{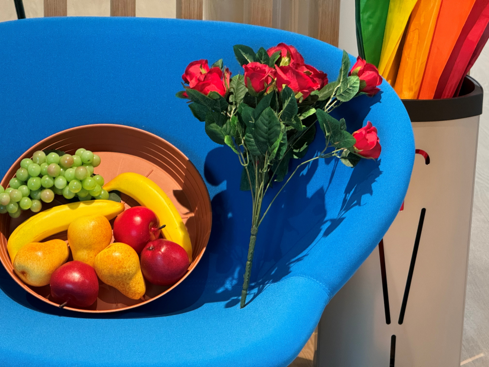} &
  \includegraphics[width=\linewidth]{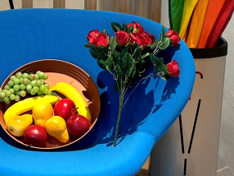} &
  \includegraphics[width=\linewidth]{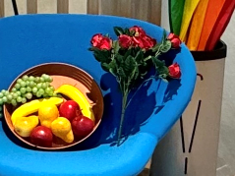} \\ \hline
  \rotatebox{90}{\centering~~~~~~~~~~~~~Pura70 Ultra} & 
  \includegraphics[width=\linewidth]{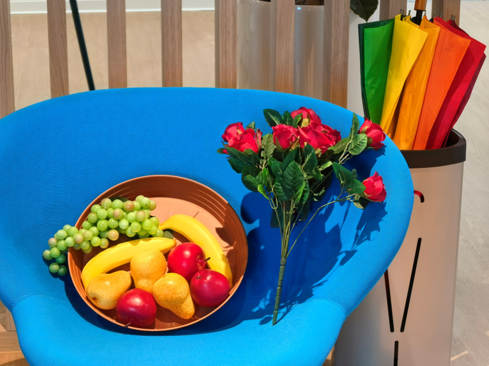} &
  \includegraphics[width=\linewidth]{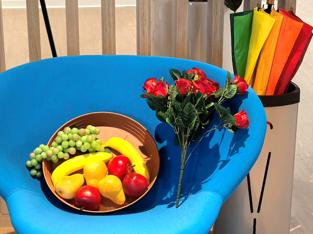} &
  \includegraphics[width=\linewidth]{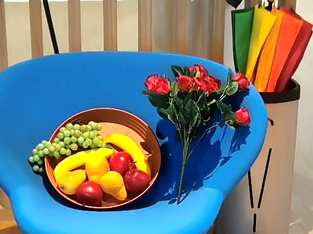} \\ \hline
  \rotatebox{90}{\centering~~~~~~~~~~~~~ NPM (Ours)} & 
  \includegraphics[width=\linewidth]{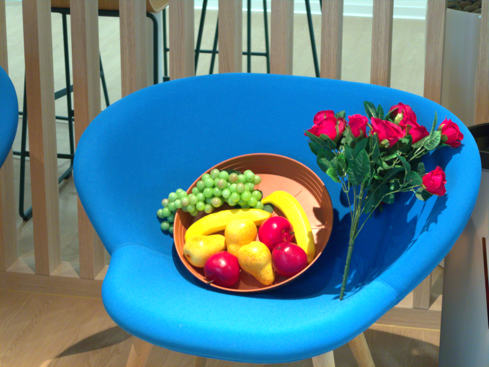} &
  \includegraphics[width=\linewidth]{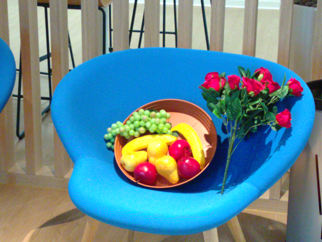} &
  \includegraphics[width=\linewidth]{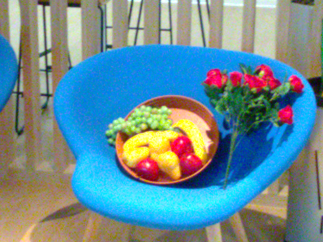} \\
  & Tele & Main & Wide \\
\end{tblr}
\caption*{
\generictext
In this example we can see some slight discrepancies in color for both the Pura70 and the iPhone between wide and main images, by looking at the scene behind the chair.
}
\label{tab:qualitative_comparisons_7}
\end{table*}

\begin{table*}[ht]
\centering
\caption*{Example 7}
\vspace{-8pt}
\begin{tblr}{
  colspec={m{0.5cm}@{}X[c]@{}X[c]@{}X[c]@{}}, 
  width = \proportion\linewidth,
  rowspec={},          
  rowsep = 2pt,
  stretch = 0,
}
  \rotatebox{90}{\centering~~~~~~~iPhone 15 Pro Max} & 
  \includegraphics[width=\linewidth]{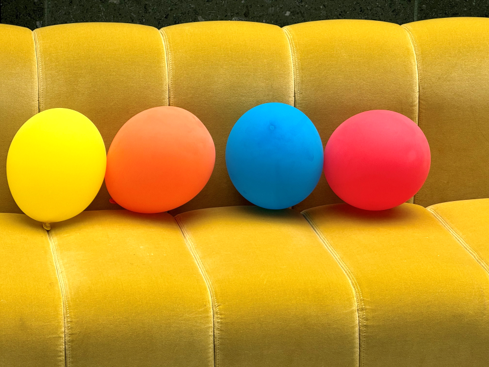} &
  \includegraphics[width=\linewidth]{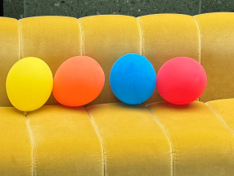} &
  \includegraphics[width=\linewidth]{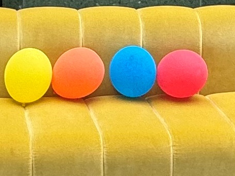} \\ \hline
  \rotatebox{90}{\centering~~~~~~~~~~~~~Pura70 Ultra} & 
  \includegraphics[width=\linewidth]{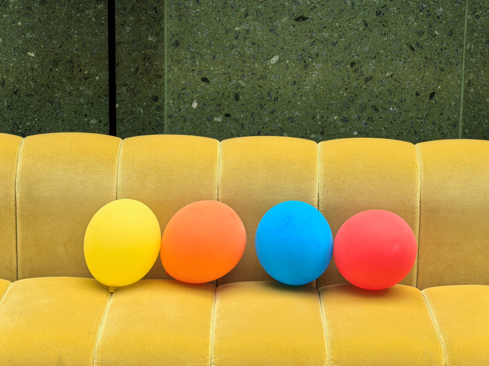} &
  \includegraphics[width=\linewidth]{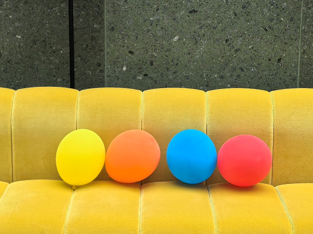} &
  \includegraphics[width=\linewidth]{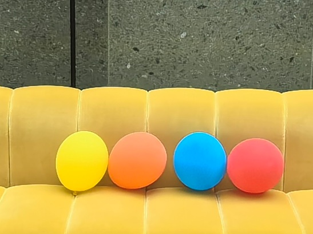} \\ \hline
  \rotatebox{90}{\centering~~~~~~~~~~~~~ NPM (Ours)} & 
  \includegraphics[width=\linewidth]{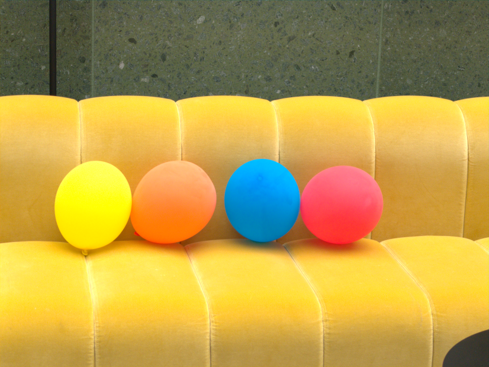} &
  \includegraphics[width=\linewidth]{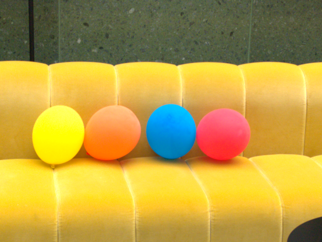} &
  \includegraphics[width=\linewidth]{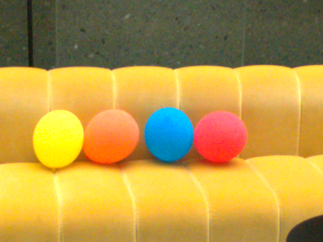} \\
  & Tele & Main & Wide \\
\end{tblr}
\caption*{
\generictext
In this example we see a simple illumination scene where all phones and our algorithm seem to perform relatively well. There are differences in intensities between the individual cameras, for example from tele to main, as well as a slight color difference for the same cameras in the Pura70.
}
\label{tab:qualitative_comparisons_9}
\end{table*}

\begin{table*}[ht]
\centering
\caption*{Example 8}
\vspace{-8pt}
\begin{tblr}{
  colspec={m{0.5cm}@{}X[c]@{}X[c]@{}X[c]@{}}, 
  width = \proportion\linewidth,
  rowspec={},          
  rowsep = 2pt,
  stretch = 0,
}
  \rotatebox{90}{\centering~~~~~~~iPhone 15 Pro Max} & 
  \includegraphics[width=\linewidth]{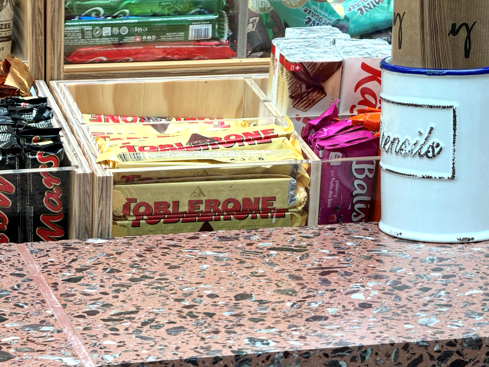} &
  \includegraphics[width=\linewidth]{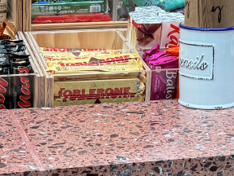} &
  \includegraphics[width=\linewidth]{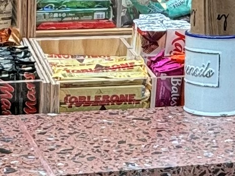} \\ \hline
  \rotatebox{90}{\centering~~~~~~~~~~~~~Pura70 Ultra} & 
  \includegraphics[width=\linewidth]{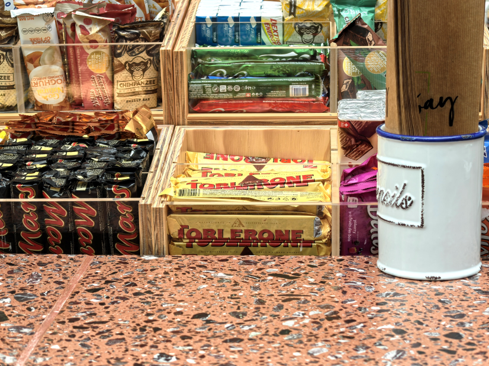} &
  \includegraphics[width=\linewidth]{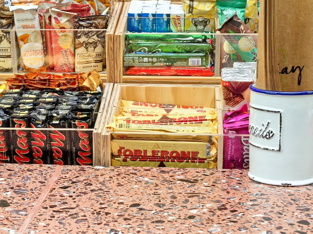} &
  \includegraphics[width=\linewidth]{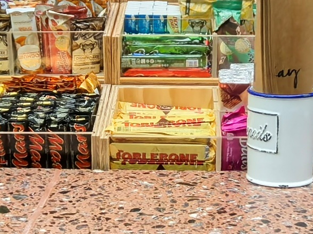} \\ \hline
  \rotatebox{90}{\centering~~~~~~~~~~~~~ NPM (Ours)} & 
  \includegraphics[width=\linewidth]{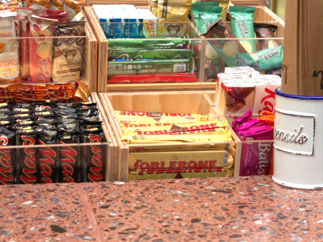} &
  \includegraphics[width=\linewidth]{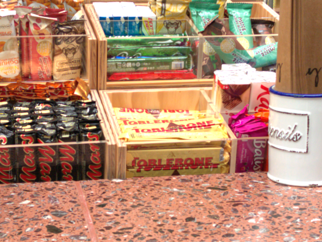} &
  \includegraphics[width=\linewidth]{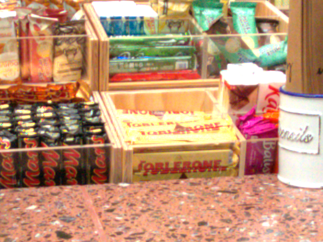} \\
  & Tele & Main & Wide \\
\end{tblr}
\caption*{
\generictext
In this example we see a dual illumination scene, where a lamp on top of the snacks is aimed directly down at them and a general illumination from outside dominates the rest of the scene. We can see a large color difference from the iPhone tele camera to the main camera, as well as slight difference in the Pura70.
}
\label{tab:qualitative_comparisons_10}
\end{table*}


\end{document}